\pdfoutput=1
\documentclass{article}
\usepackage{spconf,amsmath,amssymb,graphicx,booktabs,multirow,xcolor}
\usepackage{balance}
\usepackage{eso-pic}
\AddToShipoutPictureBG*{\AtPageUpperLeft{\raisebox{-0.62in}{\makebox[\paperwidth]{\parbox{6.6in}{\centering\footnotesize This work has been submitted to the IEEE for possible publication. Copyright may be transferred without notice, after which this version may no longer be accessible.}}}}}
\newcommand{\corr}{\mathrm{corr}}

\title{Beyond Mean Attention: Diversity-Aware, Layer-Wise Scoring for KV Cache Eviction}

\twoauthors
  {Tianfang Xie}
  {Georgia Institute of Technology\\Atlanta, GA 30332, USA\\tianfang.xie@gatech.edu}
  {Wei Zhu}
  {Zhangjiang Lab\\Shanghai, China\\15210830735@163.com}

\begin{document}
\maketitle

\begin{abstract}
KV cache eviction methods such as SnapKV and PyramidKV rank tokens solely by \emph{mean} attention over a small observation window. We study a unified score, $\mu_i+\lambda_1\sigma_i+\lambda_2\,\corr(i,S)$, adding attention \emph{dispersion} across window queries and \emph{redundancy} relative to selected tokens. For $\lambda_2<0$, the score penalizes similarity to selected tokens as in maximal marginal relevance (MMR), without extra forward passes. To test whether this relevance--diversity balance should vary with depth, we compare fixed global coefficients with three-segment and quadratic profiles. Only these depth profiles are searched on a development split under a $\sinh$ reparameterization. On all 16 English LongBench datasets with Mistral-7B at a budget of 64 entries per layer, a single global diversification constant improves 13 of 16 datasets (macro $+1.1$); the gain holds at budget 32 and narrows at 128. Per-dataset search finds no detectable layer structure on most datasets; on passage retrieval it finds a large one: a mid-layer sign flip that rewards similarity and is worth $+9.6$ over the baseline at budget 64 and, without re-tuning, $+13.2$ over the global constant at budget 128. Ablations attribute the gain to the redundancy term; replaying every accepted search state on the held-out test set separates genuine structure from tuning noise.
\end{abstract}


\section{Introduction}
\label{sec:intro}

Serving long-context large language models (LLMs) is dominated by the memory footprint of the key--value (KV) cache, which grows linearly with context length and often exceeds the model weights. Token eviction is a popular remedy: after prefill, each layer keeps only a small budget of $B$ cached entries. State-of-the-art eviction methods such as SnapKV~\cite{snapkv} and PyramidKV~\cite{pyramidkv} score each historical token by the \emph{mean} attention it receives from an observation window of the last few queries, and keep the top-$B$.

The mean, however, ignores two signals. First, window queries may agree or disagree about a token; the mean discards this \emph{dispersion}. Second, and more importantly, top-mean tokens are often near-duplicates of one another. At tight budgets, the cache then fills with \emph{redundant} copies of one salient region while complementary evidence is evicted, the problem that maximal marginal relevance (MMR) addresses in information retrieval~\cite{mmr}. Inspired by mean--variance portfolio selection~\cite{markowitz}, we treat cache eviction as building a portfolio of tokens and study the unified score $\mu_i+\lambda_1\sigma_i+\lambda_2\,\corr(i,S)$ with layer-wise coefficients.

Our contributions are threefold. (i) A diversified, layer-wise scoring framework for KV eviction that generalizes the attention-based token scoring of SnapKV/PyramidKV under a fixed budget allocation and costs no extra forward passes (Sec.~\ref{sec:method}). (ii) A comparison of a fixed global constant with two searched layer-wise parameterizations, Segments and Curve, under a random-split protocol in which the test set never influences any accepted search move (Sec.~\ref{sec:proto}). (iii) A study on all 16 English LongBench datasets that shows a fixed global setting improving 13 of 16 datasets at $B{=}64$ without per-dataset search, and one large, transferable task-specific structure, together with ablations of the score terms, results for the global constant on a second model, and a \emph{trajectory-replay} diagnostic that separates genuine structure from tuning noise (Sec.~\ref{sec:results}).

\section{Related Work}
\label{sec:related}

\noindent\textbf{KV cache compression.} StreamingLLM~\cite{streamingllm} keeps attention sinks plus a sliding window; H2O~\cite{h2o} and Scissorhands~\cite{scissorhands} evict by accumulated attention during decoding; FastGen~\cite{fastgen} selects a per-head compression policy by profiling. Closest to us, SnapKV~\cite{snapkv} scores prompt tokens by pooled observation-window attention at the end of prefill, and PyramidKV~\cite{pyramidkv} adds a depth-decreasing budget schedule. These rank tokens \emph{independently} by (a variant of) mean attention; our score contains theirs as its $\lambda_1{=}\lambda_2{=}0$ point. Redundancy-aware compression is a recent and active line: R-KV~\cite{rkv} combines attention importance with key-similarity redundancy for decoding-time compression; MixKV~\cite{mixkv} balances importance and diversity per head, relative to the global key distribution, for vision-language caches; and the concurrent TwinKV~\cite{twinkv} adds a repair pass that swaps retained near-duplicate keys for evicted orphans. We differ on three axes: redundancy is \emph{conditioned on the set already retained} through a greedy MMR rule, its weight is a searched \emph{function of depth}, and the resulting profiles are tested for transfer across budgets.

\noindent\textbf{Diversified subset selection.} Selecting items that are individually relevant yet mutually complementary is classical: MMR~\cite{mmr} greedily penalizes similarity to the selected set, and determinantal point processes~\cite{dpp} model diversity probabilistically; mean--variance portfolio theory~\cite{markowitz} makes the same trade-off for assets. We bring this view to KV eviction with a depth-dependent trade-off.

\section{Method}
\label{sec:method}

\subsection{Preliminaries}

At the end of prefill, SnapKV/PyramidKV use an observation window
$W$ comprising the last $|W|{=}8$ queries. For a layer with cache
budget $B$, token $i$ is ranked by its mean attention, with the
baselines' max-pooling retained in the scoring pipeline:
\begin{equation}
\mathrm{score}(i)=\mu_i
=\frac{1}{|W|}\sum_{q\in W}A_{q,i}.
\label{eq:mu}
\end{equation}
Here, $A_{q,i}$ denotes the attention received by token $i$ from
query $q$. The top-$B$ tokens are retained. PyramidKV decreases
budgets with depth while preserving the total budget of uniform
allocation.

\subsection{Diversity-Aware Scoring}

Our \emph{diversity-aware, layer-wise scoring for KV cache eviction}
extends mean-attention ranking to capture query disagreement and
redundancy among retained tokens. Inspired by mean--variance
portfolio selection~\cite{markowitz}, we combine individual relevance
with dispersion and set-dependent similarity. This analogy motivates
the scoring components rather than a direct portfolio objective:
attention statistics characterize each candidate, whereas key
similarity relates it to the selected set.

\noindent\textbf{Unified score.}
For a candidate $i$ and the selected token set $S$, we define
\begin{equation}
\mathrm{score}(i)
=\mu_i+\lambda_1\sigma_i+\lambda_2\,\corr(i,S),
\label{eq:score}
\end{equation}
where attention dispersion is
\begin{equation}
\sigma_i=
\left(
\frac{1}{|W|-1}\sum_{q\in W}(A_{q,i}-\mu_i)^2
\right)^{1/2},
\label{eq:sigma}
\end{equation}
and redundancy is measured by
\begin{equation}
\corr(i,S)=\max\Bigl(0,\;\max_{j\in S}\cos(k_i,k_j)\Bigr),\qquad \corr(i,\emptyset)=0.
\label{eq:corr}
\end{equation}
Here, $k_i$ is the cached key vector for token $i$.
The sample standard deviation $\sigma_i$ measures disagreement
across window queries; $\corr(i,S)$ measures the candidate's greatest
similarity to a retained token, with negative similarities clamped
to zero. Thus, $\lambda_2<0$ gives an
MMR-style penalty~\cite{mmr}, favoring complementary evidence,
whereas $\lambda_2>0$ rewards similarity. Because attention
magnitudes vary substantially across heads but cosine similarity
lies in $[-1,1]$, we normalize the base score
$s_i=\mu_i+\lambda_1\sigma_i$ per head before greedy selection:
\begin{equation}
\begin{aligned}
\tilde{s}_i
&=\frac{s_i}{\max_j s_j},\\
S
&\leftarrow S\cup
\left\{
\arg\max_{i\notin S}
\bigl(\tilde{s}_i+\lambda_2\,\corr(i,S)\bigr)
\right\}.
\end{aligned}
\label{eq:greedy}
\end{equation}
The denominator is clamped to a positive value, preserving the
base-score ranking; $\tilde{s}_i\in[0,1]$ when $\lambda_1\ge0$.
The first token maximizes $\tilde{s}_i$, and subsequent admissions
follow Eq.~\eqref{eq:greedy} until $|S|=B$.
Similarity uses keys as cached, after rotary position embedding
(RoPE), and therefore reflects both content and relative position.
Setting $\lambda_1{=}\lambda_2{=}0$ recovers the baseline ranking.
The procedure uses existing attention and keys, requiring no
additional forward passes; with candidates restricted to the top
$4B$ tokens by base score, the $B$ admissions cost $O(B^2 d)$ per head.

\subsection{Layer-Wise Parameterization}
\label{sec:proto}

The coefficients $\lambda_1$ and $\lambda_2$ control complementary
aspects of selection and need not be shared across depth.
Layer-dependent representations and PyramidKV's nonuniform budget
allocation motivate depth-varying coefficients, but limited
development data favor compact profiles over independent tuning
at every layer. Thus, we represent the coefficients with compact depth profiles.

\noindent\textbf{Depth profiles.}
For $L{=}32$ layers, independently tuning both coefficients requires
$2L$ parameters and risks overfitting 100 development questions.
We instead compare three families for the depth profile $a(\ell)$:
\emph{Global}, a shared constant; \emph{Segments}, separate constants
for shallow, middle, and deep thirds; and \emph{Curve}, a quadratic:
\begin{equation}
\begin{aligned}
a(\ell)
&=c_0+c_1t+c_2t^2,
\qquad t=\tfrac{\ell}{L-1},\\
\lambda^{(\ell)}
&=\sinh\!\bigl(a(\ell)\bigr).
\end{aligned}
\label{eq:param}
\end{equation}
Layers are indexed by $\ell=0,\ldots,L-1$, and the mapping to
$\lambda^{(\ell)}$ applies separately to $\lambda_1$ and $\lambda_2$
in every family. Segments and Curve each use three parameters per
coefficient and include Global as a special case.
For Curve, $c_0$, $c_1$, and $c_2$ control the latent profile's
height, tilt, and curvature, allowing at most one turning point.
The $\sinh$ mapping is unbounded and sign-symmetric, approximately
identity near zero, and satisfies $\cosh a\ge1$.
It also spans $\lambda\in[-3.6,3.6]$ approximately over
$a\in[-2,2]$.

\noindent\textbf{Search protocol.}
For each dataset, a fixed-seed random split yields a development
set of 100 questions (250 for lcc and repobench-p) and a held-out
test set of the remaining questions.
Only development scores determine accepted search moves; test
scores serve reporting and trajectory replay
(Sec.~\ref{sec:dynamics}).
Because $\lambda_1$ reshapes base scores and $\lambda_2$ changes
the resulting greedy selections, we optimize them jointly by
first-improvement coordinate search over the six parameters of
Segments or Curve; Global keeps the fixed setting
$\lambda_1{=}0$, $\lambda_2{=}-0.15$ and is not searched.
This setting is also the warm start of every search.
A pilot comparing
$\lambda_2\in\{-0.15,-0.3,-0.6\}$ on qasper, hotpotqa, 2wikimqa,
and passage retrieval at $B{=}64$ fixed this constant, which is
reused across all 16 datasets, budgets, and both models.
Each search allows 14 evaluations, perturbing coordinates by
$\pm0.3$ in $a$-space.
A move is accepted only when its development-score improvement
exceeds $\tau$.
Near zero, a $0.3$ perturbation in $a$ corresponds approximately
to a $0.3$ change in $\lambda$; for the redundancy coefficient,
this can shift a candidate's normalized score by about $0.3$
because $\cos\in[-1,1]$.
Larger steps help expose effects within the limited evaluation
budget, although their effective magnitude varies with $a$.
With per-question score standard deviation ${\approx}35$, the
development standard error is ${\approx}3.5$ points.
We use an acceptance tolerance of $\tau{=}0.1$ points;
this is not a statistical significance threshold.

\section{Experiments}
\label{sec:results}

\subsection{Experimental Setup}

\noindent \textbf{Datasets} We use LongBench \cite{bai2024longbench} to assess the performance of our method on tasks
involving long-context inputs. Its 16 English datasets cover single- and multi-document question
answering, summarization, few-shot learning, synthetic retrieval and counting, and code
completion.

\noindent \textbf{Experiment settings} We use Mistral-7B-Instruct-v0.2~\cite{mistral} (fp16, SDPA) in the PyramidKV implementation with a budget of $B{=}64$ entries per layer (${\approx}1\%$ of a typical prompt) unless stated otherwise, and all 16 English LongBench~\cite{bai2024longbench} datasets. Test scores use the official LongBench scorer, which truncates predictions to their first line on trec, triviaqa and samsum; search-time dev scores use raw outputs. All comparisons share the same total cache budget. The full study took ${\approx}2500$ search trials and ${\approx}260$ GPU-hours on RTX~4090s.

\begin{table}[t]
\centering
\setlength{\tabcolsep}{5pt}
\resizebox{0.46\textwidth}{!}{
\renewcommand{\arraystretch}{1.00}
\begin{tabular}{lcccc}
\toprule

\multirow{2}{*}{\textbf{Dataset}}  &   \multirow{2}{*}{\textbf{PyramidKV}}  & \multicolumn{3}{c}{\textbf{Ours}}   \\
 &  & Global & Seg. & Curve \\
\midrule
qasper & 23.73 & \textbf{25.12} & \textbf{25.12} & \textbf{25.12} \\
hotpotqa & 33.67 & \textbf{36.94} & \textbf{36.94} & 36.17 \\
2wikimqa & 23.52 & 23.33 & \textbf{23.76} & 23.12 \\
passage-retrieval & 60.60 & 66.81 & \textbf{70.17} & 65.87 \\
musique & \textbf{13.13} & 11.88 & 12.07 & 12.00 \\
narrativeqa & 18.10 & 18.42 & \textbf{18.85} & 18.42 \\
multifieldqa & 32.14 & 33.93 & \textbf{35.05} & 33.93 \\
trec & 55.00 & 56.00 & \textbf{61.00} & 56.00 \\
triviaqa & 84.53 & 85.21 & \textbf{85.33} & 84.49 \\
samsum & 33.50 & 35.23 & \textbf{35.48} & 35.23 \\
passage-count & 4.18 & \textbf{5.61} & \textbf{5.61} & \textbf{5.61} \\
lcc & \textbf{49.88} & 49.51 & 49.51 & 49.51 \\
repobench-p & 41.51 & \textbf{41.96} & \textbf{41.96} & \textbf{41.96} \\
gov-report & 18.42 & 19.08 & \textbf{19.09} & 19.08 \\
qmsum & 20.90 & 20.97 & 21.38 & \textbf{21.65} \\
multi-news & 20.17 & \textbf{21.04} & \textbf{21.04} & \textbf{21.04} \\
\midrule
Macro & 33.31 & 34.44 & \textbf{35.15} & 34.33 \\
Wins & 2 & 0 & \textbf{13} & 1 \\
\bottomrule
\end{tabular}}
\caption{Test scores (Mistral-7B, budget 64, official LongBench scorer). Bold: row best. Wins: number of datasets on which the column is the row best (ties credited to Seg.).}
\label{tab:main}
\end{table}

\subsection{Main results}

Table~\ref{tab:main} demonstrates the effectiveness of our
diversity-aware, layer-wise scoring for KV cache eviction at
$B{=}64$. With fixed coefficients $\lambda_1{=}0$ and
$\lambda_2{=}-0.15$, Global improves over PyramidKV on 13 of
16 datasets, increasing the macro score from 33.31 to 34.44
($+1.13$ points). Dataset-specific search with Segments reaches
35.15, whereas Curve obtains 34.33, indicating that greater
profile flexibility does not consistently improve performance.
The strongest task-specific benefit occurs on passage retrieval:
Segments achieves 70.17, exceeding PyramidKV by 9.57 points
and Global by 3.36 points. These results support fixed
diversification as an effective default, with additional
benefits from depth-dependent scoring on selected tasks.

\begin{table}[t]
\centering
\setlength{\tabcolsep}{4pt}
\resizebox{0.36\textwidth}{!}{
\renewcommand{\arraystretch}{1.00}
\begin{tabular}{lccc}
\toprule
Method & $B{=}32$ & $B{=}64$ & $B{=}128$ \\
\midrule
PyramidKV & 30.48 & 33.31 & 35.79 \\
Global & 31.21 & 34.44 & 35.99 \\
\textbf{Seg.} & \textbf{31.49} & \textbf{35.15} & \textbf{36.39} \\

\bottomrule
\end{tabular}}
\caption{Budget sweep on test sets (Mistral-7B). We report the macro average performance; Seg.\ profiles are deployed at $B{=}64$ and transferred unchanged.}
\label{tab:sweep}
\end{table}

\begin{figure*}[t]
\centering
\includegraphics[width=\textwidth]{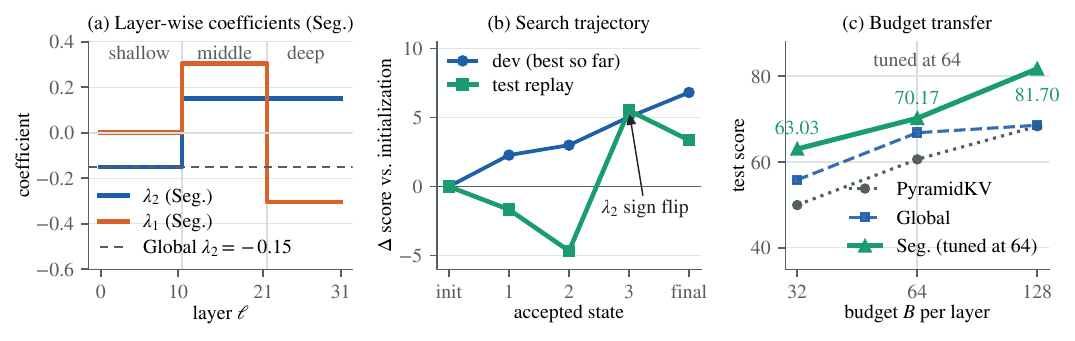}
\caption{Passage retrieval, Mistral-7B. (a) Layer-wise coefficients of the Seg.\ profile searched at $B{=}64$; dashed line: Global $\lambda_2{=}-0.15$; vertical lines: segment boundaries. (b) Development best-so-far and test replay of every accepted state of the search at $B{=}64$ (14 evaluations, $\tau{=}0.1$), as change from the initialization; only development scores determine acceptance. (c) Test score vs.\ budget: the Seg.\ profile tuned at $B{=}64$ is applied unchanged at $B{=}32$ and $128$.}
\label{fig:analysis}
\vspace{-8pt}
\end{figure*}

Table~\ref{tab:sweep} varies the budget with the profiles deployed
at $B{=}64$ transferred unchanged. Global improves
the macro score over PyramidKV by 0.73, 1.13, and 0.20 points at
$B{=}32$, 64, and 128, and Seg.\ by 1.01, 1.84, and 0.60 points,
respectively. The passage-retrieval profile transfers unchanged
(Fig.~\ref{fig:analysis}c): it scores 63.03 at $B{=}32$ and 81.70 at
$B{=}128$, 13.40 above PyramidKV and 13.16 above Global at the latter
budget.

\subsection{Ablation studies and further analysis}
\label{sec:dynamics}

\noindent \textbf{Visualization of passage retrieval results.}
Fig.~\ref{fig:analysis} shows the passage-retrieval profile (a), its search trajectory (b), and its budget transfer (c). The profile keeps $\lambda_2<0$ in shallow layers but turns it positive in the middle and deep layers. In the replay, the two accepted $\lambda_1$ updates do not improve the test score, whereas the $\lambda_2$ sign flip yields a large gain, and the profile tuned at $B{=}64$ exceeds both PyramidKV and Global at every budget.

\begin{table}[t]
\centering
\setlength{\tabcolsep}{6pt}
\resizebox{0.42\textwidth}{!}{
\renewcommand{\arraystretch}{1.05}
\begin{tabular}{lcc}
\toprule
Scoring rule & Macro & Wins \\
\midrule
Mean (baseline) & 33.31 & --- \\
Mean \& dispersion & 33.50 & 8/16 \\
Mean \& Corr & 34.57 & 13/16 \\
dispersion \& Corr (fixed $\lambda$) & 33.16 & 9/16 \\
All three terms (ours, 3 segments) & \textbf{35.15} & \textbf{14/16} \\
\midrule
All three terms, 2 segments & 34.75 & 14/16 \\
All three terms, 8 segments & 35.03 & 14/16 \\

\bottomrule
\end{tabular}}
\caption{Ablation studies of our framework (Mistral-7B, budget 64), with the Seg.\ strategy unless noted. We report the macro average scores (Macro) and the number of datasets improved over the baseline (Wins).}
\label{tab:abl}
\end{table}

\noindent \textbf{Ablation study of our framework.}
Table~\ref{tab:abl} removes one score term at a time and varies the
depth granularity. Mean attention is the backbone: a fixed rule
without it (dispersion \& Corr, $\lambda_1{=}0.3$, $\lambda_2{=}-0.15$,
no search) scores 33.16, below the 33.31 baseline, and drops passage
retrieval to 57.41. The redundancy term carries most of the gain:
searching $\lambda_2$ alone (Mean \& Corr) reaches 34.57 and improves
13 datasets, whereas searching $\lambda_1$ alone (Mean \& dispersion)
reaches 33.50 and improves 8. Searching both reaches 35.15; on
passage retrieval the joint search scores 70.17 against 66.62 for
$\lambda_2$ alone in this round. Three segments score highest (35.15); eight segments (35.03; 16
parameters, 34 evaluations) and two (34.75) stay within the noise
floor of it, so the segment count is not a sensitive choice.

\noindent \textbf{Evaluation on other LLM backbones.}
To check that the fixed rule transfers across models, we apply Global
($\lambda_1{=}0$, $\lambda_2{=}-0.15$, no re-tuning) to
Llama-3.1-8B-Instruct at $B{=}64$. qasper improves by $+5.8$
($24.4\to30.1$) and hotpotqa by $+2.0$ ($51.1\to53.1$), while the
other datasets stay within the noise floor; passage retrieval is
already at $99.0$ on this model, leaving limited headroom for a depth
profile. Per-dataset search on this model is left for future work.

\noindent \textbf{Discussion on efficiency and search cost.} The greedy admission runs once per prompt at the end
of prefill and adds no forward passes. On 12 HotpotQA prompts (mean
15.6k tokens, RTX~4090) selection takes 222\,ms per prompt versus
32\,ms for top-$k$, increasing end-to-end latency for prefill plus
32-token generation by about 14\% (2.52\,s $\to$ 2.87\,s); the
overhead comes from performing the $B$ admissions sequentially in each
layer. Global needs no per-dataset search. A
three-segment profile costs 14 development evaluations per dataset,
about 1--3 GPU-hours on one RTX~4090, and transfers across budgets
without re-tuning (Table~\ref{tab:sweep}). 

\noindent \textbf{Limitations.} Scores
are single-run on one benchmark suite with test sets of 50--250
questions per dataset, so differences of a few points on a single
dataset should be interpreted with caution; greedy search is
path-dependent, and per-dataset profiles need labeled development
data.

\section{Conclusion}
\label{sec:conclusion}

Diversified, layer-wise scoring generalizes mean-attention token scoring for KV eviction without additional forward passes. On Mistral-7B a single global diversification constant is a robust default at small cache budgets, and one task, passage retrieval, rewards a layer-wise profile with a large gain that survives changes of budget. Ablations place the benefit in the redundancy term computed on cached keys, and a trajectory-replay diagnostic makes such findings verifiable. Next steps are instance-adaptive coefficients.

\clearpage

\section{Acknowledgments}
This work was self-funded, with no
external funding. The authors declare no conflicts of interest. GPT (OpenAI) was used to polish text and review code. The authors checked all AI-assisted content and
take full responsibility for it.

\section{Compliance with Ethical Standards}
This computational study used the publicly available LongBench
benchmark~\cite{bai2024longbench}. No new data were collected, and no human
or animal subjects were involved; therefore, ethical approval was not
required.

\bibliographystyle{IEEEbib}
\bibliography{references}

\end{document}